# Tract orientation mapping for bundle-specific tractography

Jakob Wasserthal[1,2], Peter F. Neher[1], and Klaus H. Maier-Hein[1,3,⋆]

[1] Division of Medical Image Computing, German Cancer Research Center (DKFZ), Heidelberg, Germany
[2] Medical Faculty, Heidelberg University, Germany
[3] Section for Automated Image Analysis, Heidelberg University Hospital, Germany

**Abstract.** While the major white matter tracts are of great interest to numerous studies in neuroscience and medicine, their manual dissection in larger cohorts from diffusion MRI tractograms is time-consuming, requires expert knowledge and is hard to reproduce. Tract orientation mapping (TOM) is a novel concept that facilitates bundle-specific tractography based on a learned mapping from the original fiber orientation distribution function (fODF) peaks to a list of tract orientation maps (also abbr. TOM). Each TOM represents one of the known tracts with each voxel containing no more than one orientation vector. TOMs can act as a prior or even as direct input for tractography. We use an encoder-decoder fully-convolutional neural network architecture to learn the required mapping. In comparison to previous concepts for the reconstruction of specific bundles, the presented one avoids various cumbersome processing steps like whole brain tractography, atlas registration or clustering. We compare it to four state of the art bundle recognition methods on 20 different bundles in a total of 105 subjects from the Human Connectome Project. Results are anatomically convincing even for difficult tracts, while reaching low angular errors, unprecedented runtimes and top accuracy values (Dice). Our code and our data are openly available.

## 1 Introduction

Diffusion tractography would be much simpler to solve if there existed only one of the major tracts in the brain. In reality, though, multiple tracts co-exist and overlap, resulting in multiple fiber orientation distribution function (fODF) peaks per voxel and larger bottleneck situations with tracts per voxel outnumbering the peaks per voxel. In consequence, tractography is highly susceptible to false positives [5, 4]. The only safe solution around false positives today is the explicit dissection of anatomically well-known tracts. While manual dissection protocols [10] can be considered the current gold standard, a variety of approaches was already developed for automating the process: *Region-of-interest-based* approaches filter streamlines based on their spatial relation to cortical or other anatomically defined regions, which are typically transferred to subject space via atlas registration and segmentation techniques [12, 14]. *Clustering-based* approaches group and select streamlines by measuring intra- and inter-subject streamline similarities, referring to existing reference bundles in atlas space [2, 6, 7].

---

⋆ Corresponding author. E-mail address: k.maier-hein@dkfz.de

Concept-wise, many previous approaches have opted for performing a rather blind whole brain tractography and then investing the effort in streamline space, clearing the tractograms from spurious streamlines and grouping the remaining ones. We propose a novel concept, Tract Orientation Mapping (TOM), that approaches the problem before doing tractography by learning tract-specific peak images (tract orientation maps). These can be used as a prior – relating them to Rheault *et al.* [8], who employed registered atlas information as a tract-specific prior – or directly as orientation field for tractography. The larger-scale quantitative evaluation of such approaches is challenging due to the effort required to manually produce high quality reference tracts. To address this, an interactive process in form of auxiliary tools was designed in support of the expert. This helped us achieving high quality semi-automatic reference dissections of 20 bundles in a total of 105 Human Connectome Project (HCP) subjects. On basis of this novel data set, TOM was compared to several state of the art methods and was able to set new standards in terms of quality, quantity and runtime.

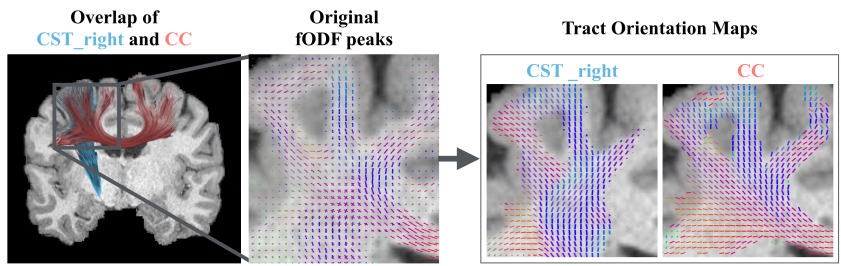

**Fig. 1.** Exemplary depiction of a slice through two of the reference tracts, the original fODF peak image and the corresponding reference TOMs (cf. Fig. 3 for abbr.).

## 2   Methods

Given a set of reference tracts, TOM is based on a learned mapping from the original fODF peak image to a list of tract orientation maps (cf. Fig. 1). Please note that TOM might refer to the general concept (tract orientation mapping) as well as to one of the tract orientation maps itself. Each TOM represents one tract, and each voxel contains one orientation vector representing the local tract orientation, i.e. the local mean streamline orientation of the corresponding reference tract. The mapping is learned via training a fully convolutional neural network (FCNN) with the original fODF peak image as input to regress the different TOMs as output channels. The network is then used to predict estimated TOMs for previously unseen subjects. These can be used as a tract-specific prior or employed directly for tractography (Fig. 2).

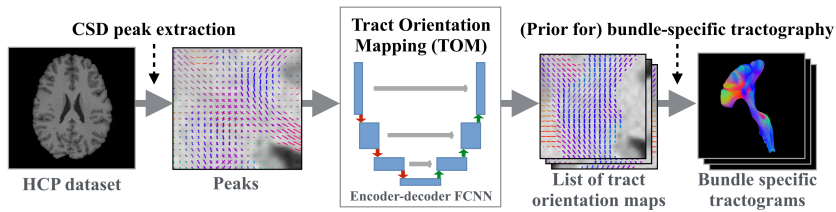

**Fig. 2.** Pipeline overview: Constrained spherical deconvolution (CSD) is applied to obtain the three principal fODF directions per voxel. An encoder-decoder FCNN maps the original fODF peak image to bundle-specific peak images, i.e. TOMs, which are then employed as a prior for or as direct input for bundle-specific tractography.

**Model** We implemented TOM using an encoder-decoder FCNN with long-range skip connections. Our FCNN uses the same number of layers and filters as the U-Net architecture [9], but with a input size of 144x144 and leaky ReLU activations. It has 9 input channels (the three principal fODF peaks) and 60 output channels (one 3D vector for each of the 20 bundles). The high dimensionality and spatial resolution of our data let us opt for a memory-efficient 2D instead of a 3D architecture. 2D slices were sampled along the y-axis. The decision of using fODF peaks as an input rather than raw image values or parametric representations of the signal, such as spherical harmonics coefficients, was also driven by the desire to reduce memory demand, with the side effect of becoming more independent of the acquisition scheme. fODF peaks were extracted using constrained spherical deconvolution (CSD) and peak extraction in MRtrix [3].

**Training** Since we are solving a regression task, we employed linear activation functions in the last layer. As loss function we used weighted cosine similarity combined with mean squared error of the $l^2$-norm

$$loss(\hat{y}, y, w) = \frac{1}{N} \sum_{i=0}^{N} \left[ -\frac{1}{w} * \left| \frac{\langle \hat{y}_i, y_i \rangle}{\|\hat{y}_i\|_2 * \|y_i\|_2} \right| + \frac{w * (\|\hat{y}_i\|_2 - \|y_i\|_2)^2}{w_{max}} \right] \quad (1)$$

with $N$ being the number of classes, $y$ the training target, $\hat{y}$ the network output and $w$ a weighting factor which was used to handle the class imbalance between background and bundles and reinforce the training signal of the bundles. We set $w = 10$ at epoch 0 and linearly reduced it to $w = 3$ at epoch 300. For $y = 0$, we set $w = 0$. We used a learning rate of 0.001, an Adamax optimizer, 300 epochs of training and a batch size of 44. The input images were normalized to zero mean and standard deviation one. In the experiments, we applied 5-fold cross validation with splits for training, validation and testing. Hyperparameters were optimized on the validation set, network parameters producing optimal Dice scores were used for testing. Dice scores were calculated by thresholding the $l^2$-norm of the peaks.

**Bundle-specific tractography** There are different flavors of bundle-specific tractography on basis of the estimated TOMs. The maps can be used for direct *TOM tractography*. Here, we applied deterministic MITK Diffusion tractography for this purpose, min. length 50mm, one seed per voxel. Alternatively, a *TOM prior* can be applied during tractography on the original data. Here, we implemented this by amplifying the bundle-specific peaks using a weighted mean between original and prior peaks, similar to [8]. Last but not least, completely sticking to the original data, a *TOM-based peak selection* might be performed for tractography on the best-matching original peaks. For all three flavors, streamlines are stopped whenever the TOM becomes zero, which is the case outside of the respective bundle.

**Reference Data** The approach was evaluated on a newly created database of 105 HCP subjects, each with a semi-automatic dissection of 20 prominent white matter tracts [13]. HCP imaging parameters were: 1.25 $mm$ isotropic resolution, 270 gradient directions, three b-values $b = [1000; 2000; 3000]\, s/mm^2$. Starting point per subject was a 10 million streamline tractogram generated using MRtrix CSD and iFOD2 (min. length: 40mm) [11]. TractQuerier [12] was then used to extract a first approximation of each bundle. Several interactive auxiliary tools were then employed and interactively combined in a workflow implemented in MITK Diffusion to achieve high quality reference dissections of each tract: (a) Manual definition of inclusion and exclusion ROIs, (b) QuickBundles [1] clustering for detection of small or spurious streamline clusters, (c) detection of streamlines that run through low fiber density voxels. Additionally, streamlines that run back and forth inside the target bundle, i.e. making 180 degree turns within less than 30mm distance, were also removed. To make the high quality of this dataset accessible, we openly published our reference tracts for all 105 subjects: https://doi.org/10.5281/zenodo.1088277

**Reference Methods** We use four state of the art automatic tract delineation methods as baseline, including clustering- and as ROI-based approaches:
*RecoBundles* [2] registers the tractogram to a reference subject and uses clustering to detect streamlines that are similar to the reference tracts. It was run with default parameters on 5 different reference subjects, aggregated by taking the mean. Using all 63 training subjects as reference subjects was computationally infeasible due to the long runtime of RecoBundles for 10 million fibers.
*WhiteMatterAnalysis* (WMA) [6,7] clusters streamlines across several subjects and produces a corresponding atlas. Each cluster in the atlas is assigned to a specific anatomical bundle. Registering new subjects to the atlas enables automated bundle delineation. We use the pretrained WMA atlas containing 800 clusters. We manually optimized the nine predefined mappings of clusters to anatomical tracts to better align it with our reference. During this process, we were inherently limited by the finite set of distinct clusters offered by the atlas. We chose not to extend the mapping to the 11 remaining reference tracts, which would require considerable effort given the amount of clusters. Applying WMA to 10 million streamlines requires >100GB of memory, producing clusters with substantial amounts of false positives. Thus, streamline counts were reduced to 500k, requiring approx. 30GB of memory and creating cleaner clusters.

*TractQuerier* [12] extracts tracts based on the regions the streamlines have to start in, end in and (not) run through defined in its own query language. We used the same queries as used in our pipeline for extracting the reference data (but without all the subsequent filtering).

*Atlas Registration* was additionally implemented as an in-house reference method. All training subjects were iteratively registered to the same space using FA-based symmetric diffeomorphic registration. The binary reference tract masks were averaged in atlas space and used for removing fibers exiting the masks in unseen registered test subjects.

## 3    Results

**Quantitative** Figure 3 shows overall mean as well as tract-specific Dice scores for each of the methods. These were calculated on basis of the binary tract masks in comparison to the reference. For the overall means, differences between TOM and all other methods were statistically significant according to Wilcoxon signed-rank tests ($p < 0.001$). The mean angular error for the peaks predicted by TOM in comparison to the reference TOM (i.e. the mean streamline orientation in the reference) was $16.7° \pm 0.5$. This error is smaller than the mean angular error between the reference TOM and the best-matching original fODF peak in each voxel ($18.8° \pm 0.5$).

**Qualitative** Figure 4 shows qualitative results for the corticospinal tract, the commissure anterior and the inferior occipito-frontal fascicle. We also show the results of RecoBundles which had the best Dice score of all reference methods in our quantitative evaluation. *TOM tractography* shows spatially very coherent and complete bundles, without spurious streamlines. RecoBundles on a deterministic tractogram, however, shows incomplete reconstructions, missing critical parts like the lateral projections of the CST. RecoBundles on a probabilistic tractogram finds the complete bundles (except for the left part of the CA), but introduces false positive and spatially incoherent streamlines.

We also evaluated how the different flavors of TOM-based bundle-specific tractography affect the reconstructions (Figure 5). *TOM-based peak selection* failed to reconstruct some difficult parts like the lateral projections of CST. Tractography with a *TOM prior* resulted in a complete CST reconstruction. Similar but more smooth results were obtained using direct *TOM tractography*. This finding is well-aligned with our above-reported observation regarding the angular errors of predicted and original peaks when compared to the reference peaks.

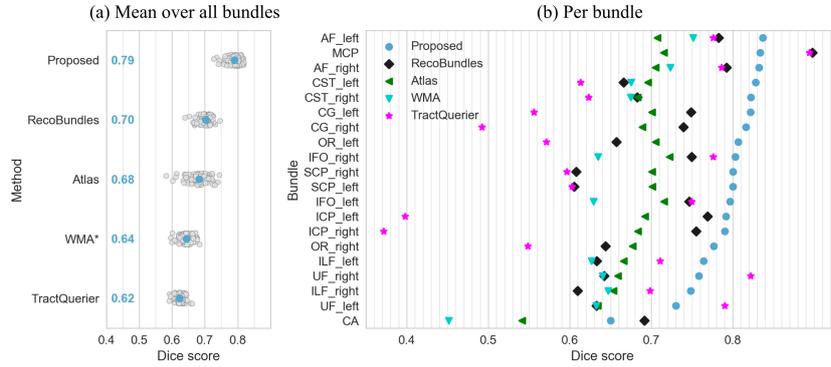

**Fig. 3.** Results of 5-fold cross validation on 105 subjects for the analyzed methods. (AF: Arcuate fascicle, CA: Commissure anterior, CST: Corticospinal tract, CG: Cingulum, ICP: Inferior cerebellar peduncle, MCP: Middle cerebellar peduncle, SCP: Superior cerebellar peduncle, ILF: Inferior longitudinal fascicle, IFO: Inferior occipito-frontal fascicle, OR: Optic radiation, UF: Uncinate fascicle) *: Mean score over the nine tracts provided by WMA analysis, see methods.

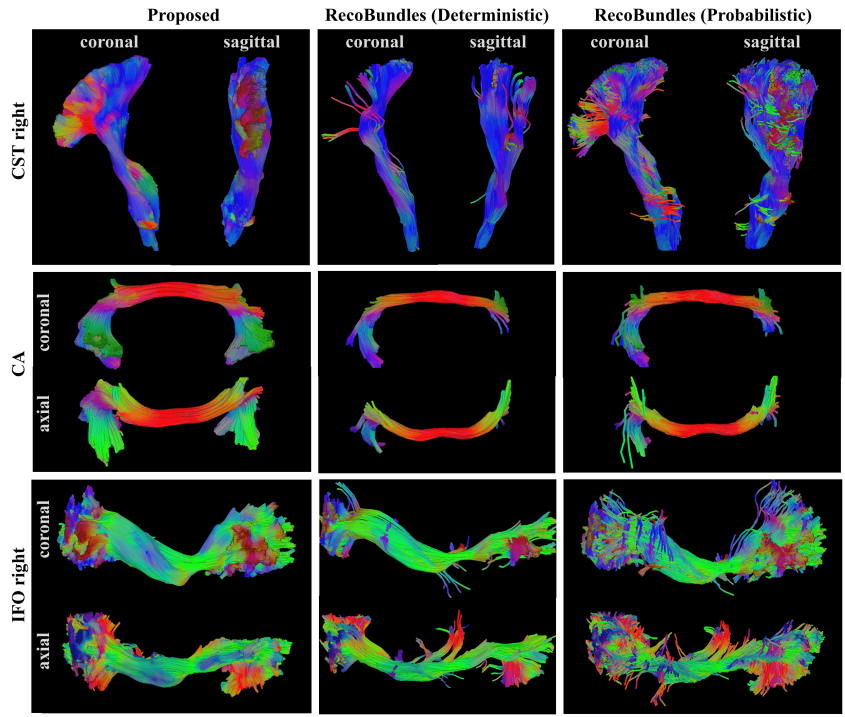

**Fig. 4.** Qualitative comparison on a random subject of the proposed method with RecoBundles, applied to deterministic and probabilistic MITK Diffusion tractography, min. length 50mm, 1 million streamlines, fODF from MRtrix multi-shell multi-tissue CSD.

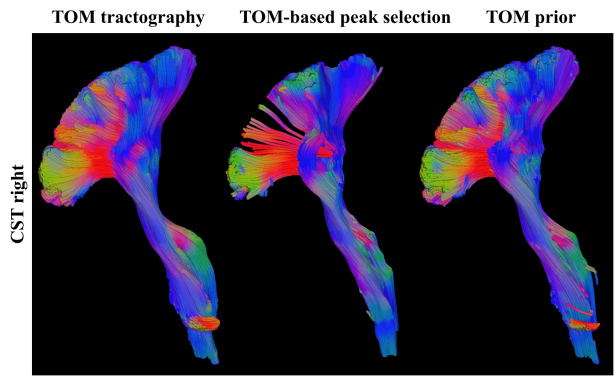

**Fig. 5.** Right CST of a random test subject reconstructed using the different variants of TOM-based bundle-specific tractography.

**Runtime** For the reconstruction of 20 bundles in one subject, TOM required less than 2 minutes, making it more than five times faster than the second fastest method Atlas Registration (approx. 11 minutes). TractQuerier, RecoBundles and WMA required much longer processing times with around 17, 97 and 938 minutes, respectively (see supplementary materials for more information).

## 4  Discussion and Conclusion

We presented a novel concept for learning-based bundle-specific tractography that employs estimated bundle-specific peak images, i.e. TOMs. The results are highly encouraging when considering quantitative, qualitative and runtime measures. One interesting finding was, in comparison to the reference, the lower angular error of the estimated TOM in comparison to the voxel-wise best-matching fODF peaks. This was reflected in the experiments by improved performances with increasing influence of the TOM. While the fODF represents a probability distribution generating many streamlines during probabilistic tractography, the TOM represents only those streamlines that were selected as a reference.

A limitation of our evaluation was the use of TractQuerier both during the creation of the reference and during validation, inducing a potential positive bias for the method. Furthermore, WMA evaluation was only available for nine out of 20 tracts, which is not necessarily comparable. The margin between WMA and the proposed method remains, though, when restricting TOM to these same nine tracts. Regarding the reference data, although we applied extensive efforts to mitigate the limitations of tractography, our tracts do not represent a real *ground truth*. They are also subject to slight variations in the detailed anatomical definition of tracts, e.g. when it comes to exact start and end regions. Despite these limitations, to the best of our knowledge, the employed data set represents one of the best existing in-vivo approximations of known white matter anatomy in a cohort of that size. The presented approach is based on supervised learning, bearing the inherent limitation of depending on the availability and quality of training data. This is similar to RecoBundles, WMA and Atlas Registration, which also require matching atlas information with defined reference tracts. We have not yet studied the limits of TOM with respect to minimizing the amount of training data. Moreover, we have not yet studied the feasibility of applying HCP-trained TOM to non-HCP datasets, potentially with the help of domain adaptation techniques. While first in-house experiments on schizophrenia patients seem promising, this needs more evaluation and remains a potentially rewarding line of research. The code of our method is available as an easy to use python package: https://github.com/MIC-DKFZ/TractSeg

## References

1. Garyfallidis et. al: QuickBundles, a Method for Tractography Simplification. Frontiers in Neuroscience 6 (2012)
2. Garyfallidis et al.: Recognition of white matter bundles using local and global streamline-based registration and clustering. NeuroImage 170, 283–295 (2017)
3. Jeurissen et al.: Multi-tissue constrained spherical deconvolution for improved analysis of multi-shell diffusion mri data. NeuroImage 103, 411–426 (2014)
4. Knösche et al.: Validation of tractography: Comparison with manganese tracing. Human Brain Mapping 36(10), 4116–4134 (2015)
5. Maier-Hein et al.: The challenge of mapping the human connectome based on diffusion tractography. Nature Communications 8(1), 1349 (2017)
6. O'Donnell et al.: Automatic tractography segmentation using a high-dimensional white matter atlas. IEEE Transactions on Medical Imaging 26(11) (2007)
7. O'Donnell et al.: Automated white matter fiber tract identification in patients with brain tumors. Neuroimage: Clinical 13, 138–153 (2017)
8. Rheault et al.: Bundle-specific tractography using voxel-wise orientation priors. In: ISMRM (2018)
9. Ronneberger et al.: U-net: Convolutional networks for biomedical image segmentation. In: MICCAI. pp. 234–241. Springer (2015)
10. Stieltjes et al.: Diffusion Tensor Imaging - Introduction and Atlas. Springer Berlin Heidelberg (2013)
11. Tournier et al.: Improved probabilistic streamlines tractography by 2nd order integration over fibre orientation distributions. In: ISMRM. p. 1670 (2010)
12. Wassermann et al.: The white matter query language: a novel approach for describing human white matter anatomy. Brain Struct. Funct. 221(9), 4705–4721 (2016)
13. Wasserthal et al.: Tractseg - fast and accurate white matter tract segmentation. arXiv:1805.07103 (2018)
14. Yendiki et al.: Automated probabilistic reconstruction of white-matter pathways in health and disease using an atlas of the underlying anatomy. Front. Neuroinform. 5(23), 12–23 (2011)